\documentclass[10pt,twocolumn,letterpaper]{article}

\usepackage{wacv}
\usepackage{times}
\usepackage{epsfig}
\usepackage{graphicx}
\usepackage{amsmath}
\usepackage{amssymb}
\usepackage[utf8]{inputenc}
\usepackage{cite}
\usepackage{wrapfig}
\usepackage{lscape}
\usepackage{rotating}
\usepackage{caption}
\usepackage{mathtools}
\usepackage{verbatim}
\usepackage{booktabs}
\usepackage{multirow}
\usepackage{bbm}
\usepackage{multicol}
\usepackage{nicefrac}
\usepackage{hyperref}

\DeclareMathOperator{\E}{\mathbb{E}}
\DeclareMathOperator*{\argmin}{arg\,min}
\DeclareMathOperator*{\argmax}{arg\,max}

\newcommand{\beq}{\begin{equation}}
\newcommand{\eeq}{\end{equation}}

\newcommand{\mr}[1]{\mathrm{#1}}

\newcommand{\Loss}{L}

\newcommand{\LossCrossEntropy}{\Loss^S_{\mr{gen}}}
\newcommand{\LossNorm}{\Loss^I_{\mr{gen}}}
\newcommand{\LossUnlabAdv}{\Loss_{\mr{disc}}}
\newcommand{\LossCycle}{\Loss_{\mr{cycle}}}
\newcommand{\LossTotal}{\Loss_{\mr{total}}}

\newcommand{\ImgLab}{x}
\newcommand{\SegLab}{y}
\newcommand{\ImgUnlab}{x'}

\newcommand{\Xentr}{\mathcal{H}}

\newcommand{\Params}{\theta}

\newcommand{\DataLab}{\mathcal{L}}
\newcommand{\DataUnlab}{\mathcal{U}}

\newcommand{\GenStoI}{G_{SI}}
\newcommand{\GenItoS}{G_{IS}}

\newcommand{\DistrSegLab}{\mathcal{Y}_{\DataLab}}
\newcommand{\DistrImgLab}{\mathcal{X}_{\DataLab}}
\newcommand{\DistrImgUnlab}{\mathcal{X}_{\DataUnlab}}

\newcommand{\visVOC}[1]{\includegraphics[width=0.147\linewidth]{Figs/VisualResults/VOC/#1}}

\newcommand{\visCity}[1]{\includegraphics[width=0.3\linewidth]{Figs/VisualResults/Cityscapes/#1}}

\newcommand{\visACDC}[1]{\includegraphics[width=0.147\linewidth]{Figs/VisualResults/ACDC/#1}}

\newcommand{\visVOCA}[1]{
\includegraphics[width=0.15\linewidth]{Figs/VisSupMat/VOC/#1.png} &
\includegraphics[width=0.15\linewidth]{Figs/VisSupMat/VOC/#1_gt.png} &
\includegraphics[width=0.15\linewidth]{Figs/VisSupMat/VOC/#1_partial.png} &
\includegraphics[width=0.15\linewidth]{Figs/VisSupMat/VOC/#1_semiSup.png}  \\}

\newcommand{\visCityA}[1]{
\includegraphics[width=0.235\linewidth]{Figs/VisSupMat/Cityscapes/#1_leftImg8bit_img.jpg} &
\includegraphics[width=0.235\linewidth]{Figs/VisSupMat/Cityscapes/#1_leftImg8bit_gt.png} &
\includegraphics[width=0.235\linewidth]{Figs/VisSupMat/Cityscapes/#1_leftImg8bit_partial.png} &
\includegraphics[width=0.235\linewidth]{Figs/VisSupMat/Cityscapes/#1_leftImg8bit_semiSup.png}  \\}

\wacvfinalcopy 

\ifwacvfinal\fi
\begin{document}


\title{Revisiting CycleGAN for semi-supervised segmentation}


\author{Arnab Kumar Mondal\\
IIT Kharagpur\\
{\tt\small sanu.arnab@gmail.com}
\and
Aniket Agarwal \\
IIT Roorkee\\
{\tt\small aagarwal@ma.iitr.ac.in}
\and
Jose Dolz \\
ETS Montreal\\
{\tt\small jose.dolz@etsmtl.ca}
\and 
Christian Desrosiers \\
ETS Montreal\\
{\tt\small christian.desrosiers@etsmtl.ca}
}

\maketitle
\ifwacvfinal\thispagestyle{empty}\fi

\begin{abstract}
In this work, we study the problem of training deep networks for semantic image segmentation using only a fraction of annotated images, which may significantly reduce human annotation efforts. Particularly, we propose a strategy that exploits the unpaired image style transfer capabilities of CycleGAN in semi-supervised segmentation. Unlike recent works using adversarial learning for semi-supervised segmentation, we enforce cycle consistency to learn a bidirectional mapping between unpaired images and segmentation masks. This adds an unsupervised regularization effect that boosts the segmentation performance when annotated data is limited. Experiments on three different public segmentation benchmarks (PASCAL VOC 2012, Cityscapes and ACDC) demonstrate the effectiveness of the proposed method. The proposed model achieves 2-4\% of improvement with respect to the baseline and outperforms recent approaches for this task, particularly in low labeled data regime. 
\end{abstract}

\section{Introduction}


Deep learning methods have recently emerged as an efficient solution for semantic image segmentation, achieving outstanding performance in a wide range of applications like analyzing natural scenes, autonomous driving or medical imaging. Despite their success, a main limitation of these methods is the need for large training datasets of pixel-level annotated images. Acquiring such labeled images is a time consuming process that may require user expertise in various scenarios. This impedes the applicability of deep models to applications where labeled images are scarce.

Semi-supervised learning (SSL) has been proposed to overcome the shortage of labeled data. In this scenario, we assume that a large set of unlabeled images is available during training, in addition to a small set of images with strong annotations. Consider a SSL segmentation setting with two distinct subsets: $\DataLab = \{(\ImgLab_i, \SegLab_i)\}_{i = 1}^n$ contains labeled images $\ImgLab_i$ and their corresponding ground-truth mask $\SegLab_i$, and $\DataUnlab = \{\ImgUnlab_{i}\}_{i = 1}^m$ contains {\em unlabeled} images $\ImgUnlab_{i}$ (typically $m \gg n$). In this setting, the objective is often formulated as maximizing a log-likelihood with respect the learning parameters $\Params$ of a deep network, through the supervision provided by the labeled set $\DataLab$. On the other hand, unsupervised images in $\DataUnlab$ can be leveraged in different ways, typically introducing a regularization effect in deep models and therefore improving their generalization capabilities. 


Generative adversarial networks (GANs) \cite{NIPS2014_5423} have shown to be an efficient solution for unsupervised domain adaptation \cite{hoffman2016fcns,hoffman2017cycada,tsai2018learning,tzeng2017adversarial}, a problem related to semi-supervised learning. GAN-based methods for domain adaptation use adversarial learning to match the distributions of source and target data, commonly at the input or in feature space. Recently, the CycleGAN model \cite{zhu2017unpaired} has become a popular choice to transfer image style between domains, as it eliminates the restriction of corresponding image pairs during training \cite{isola2017image}. This model finds a mapping between source and target images which preserves key attributes between the input and the transformed image using a cycle consistency loss. While CycleGAN has been widely employed to learn a mapping between different domains, it has not yet been investigated in more traditional semi-supervised scenarios where there is no domain shift between labeled and unlabeled data.

In this work, we leverage the unpaired domain adaptation ability of CycleGAN to learn a bidirectional mapping from unlabeled real images to available ground truth masks. This mapping, learned in conjunction with the standard supervised mapping from labeled images to their corresponding labels, acts as an unsupervised regularization loss which helps train the network when labeled data is limited. The proposed method contrasts with recent work on domain adaptation for segmentation \cite{hoffman2017cycada,jiang2018tumor}, where the CycleGAN is employed to map images across two domains. It also differs significantly from recent work using GAN-generated images for semi-supervised segmentation \cite{souly2017semi}, in which cycle consistency is not enforced. The main contributions of this paper can be summarized as follows: 
\begin{enumerate}\setlength\itemsep{0.1em}
\item  To our knowledge, this is the first semi-supervised segmentation method using CycleGAN to learn a cycle-consistent mapping between unlabeled real images and ground truth masks. The proposed technique acts as an unsupervised regularization prior which improves segmentation performance when labeled data is limited.
\item We validate our approach on three challenging segmentation tasks from different applications (i.e., natural scenes, autonomous driving and medical imaging), and show that our method is dataset-independent and effective for a wide range of scenarios.
\item Additionally, we present an ablation study which analyzes the effect of various components of the proposed unsupervised loss and demonstrates the usefulness of these components for improving performance. We believe this analysis is important for future investigations of CycleGANs applied to semi-supervised segmentation.
\end{enumerate}

The rest of the paper is organized as follows. In Section \ref{sec:related_works}, we give a brief overview of relevant work on semantic segmentation with a focus on semi-supervised learning and adversarial learning. Section \ref{sec:methodology} then presents our model which is evaluated on three challenging datasets in Section \ref{sec:experiments}. Finally, we conclude with a summary of our main contributions and results.

\begin{figure*}[ht!]
 \centering
 \includegraphics[width=0.95\linewidth]{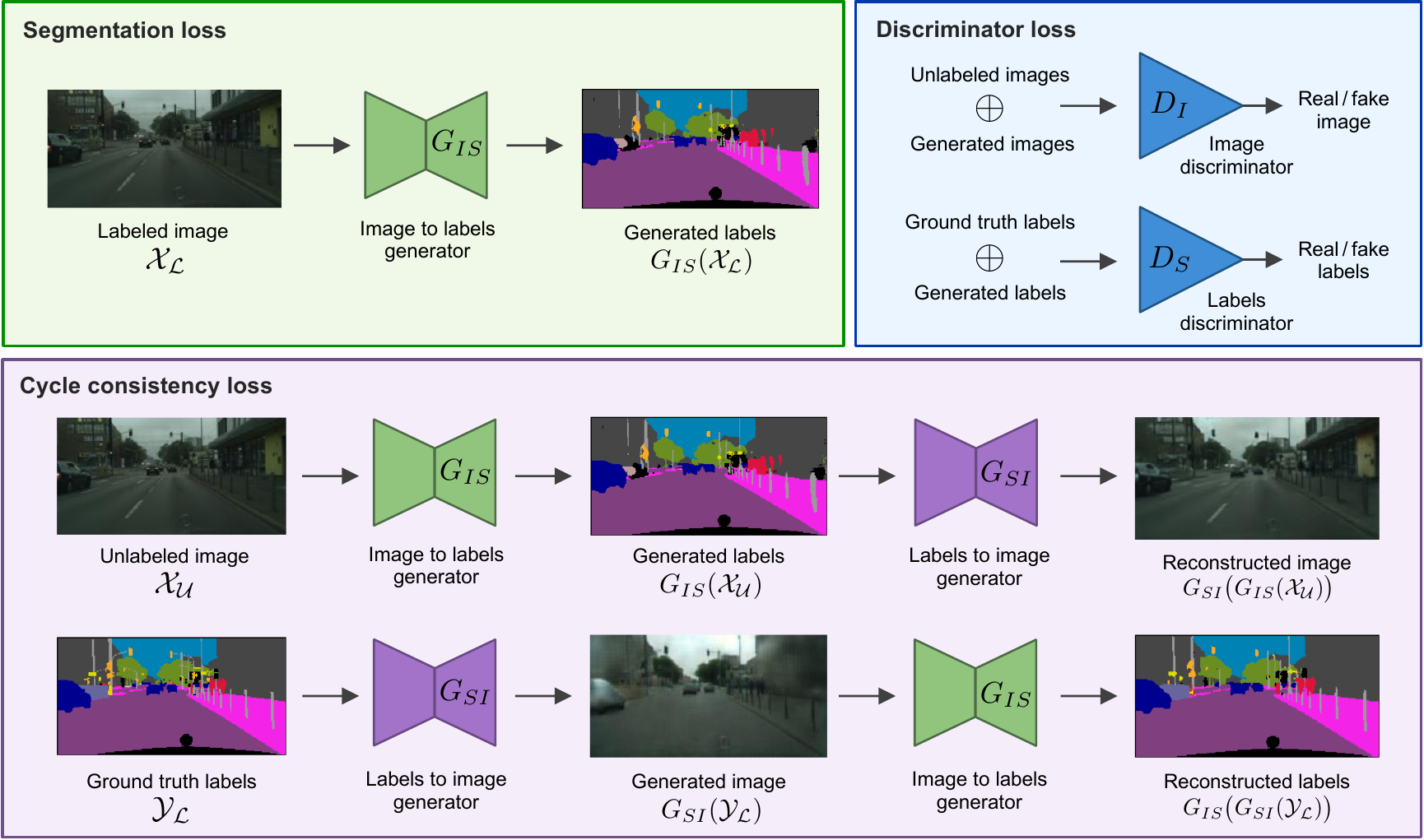} 
 \caption{Schematic explaining the working of our model. The model contains four networks which are trained simultaneously. 
 }
 \label{fig:architecture2}
\end{figure*}

\section{Related work}\label{sec:related_works}

Supervised methods based on convolutional neural networks (CNNs) are driving progress in semantic segmentation \cite{long2015fully,ronneberger2015u,Chen2018DeepLab}. 
Despite their success, training these networks requires a large number of densely-annotated images which are expensive to obtain. A solution to address this limitation is weakly-supervised learning, where easier-to-obtain annotations like image-level tags \cite{papandreou2015weakly,kervadec2019constrained,zhou2019collaborative}, bounding boxes \cite{dai2015boxsup,rajchl2017deepcut} or scribbles \cite{lin2016scribblesup} are instead used to train segmentation models. However, weakly-supervised methods still require some human interaction, which may be difficult to get in certain scenarios. 

Semi-supervision is a special type of weakly-supervised learning where many unlabeled images are also available for training \cite{bai2017semi,baur2017semi,min2018robust,perone2018unsupervised,peng2019deep,zhou2018semi}. Instead of relying on weak annotations, semi-supervised learning (SSL) typically uses domain- or task-agnostic properties of the data to regularize learning. Recently, several SSL methods have been proposed for semantic segmentation, for instance based on self-training \cite{bai2017semi}, distillation \cite{zhou2018semi}, attention learning \cite{min2018robust}, manifold embedding \cite{baur2017semi}, co-training \cite{peng2019deep}, and temporal ensembling \cite{perone2018unsupervised}. As these methods, the proposed approach can also leverage unlabeled image directly, without the need for weak annotations or task-specific priors. 

Adversarial learning has also shown great promise for training deep segmentation models with few strongly-annotated images \cite{souly2017semi,Hung_semiseg_2018,zhang2017deep}. An interesting approach to include unlabeled images during training is to add a discriminator network in the model, which must determine whether the output of the segmentation network corresponds to a labeled or unlabeled image \cite{zhang2017deep}. This encourages the segmentation network to have a similar distribution of outputs for images with and without annotations, thereby helping generalization. A potential issue with this approach is that the adversarial network can have a reverse effect, where the output for annotated images becomes growingly similar to incorrect predictions obtained for unlabeled images. A related strategy uses the discriminator to predict a confidence map for the segmentation, enforcing this output to be maximum for annotated images \cite{Hung_semiseg_2018}. For unlabeled images, areas of high confidence are used to update the segmentation network in a self-teaching manner. The main limitation of this approach is that a confidence threshold must be provided, the value of which can affect performance. 

Until now, only a single work has applied Generative Adversarial Networks (GANs) for semi-supervised segmentation \cite{souly2017semi}. In this previous work, generated images are used for training in addition to both labeled and unlabeled data. The trained segmentation network must predict the correct labels for real images or a special \emph{fake} label for generated images. For this method to work, fake images should be generated from outside the distribution of real images so that the segmentation network learns a better representation of the manifold (i.e., fake images constitute negative examples). In contrast, our method uses cycle-consistent GANs to better estimate the distribution of real images and their corresponding segmentation masks.  

\section{Methodology}\label{sec:methodology}

\subsection{CycleGAN for semi-supervised segmentation}

The proposed architecture for semi-supervised segmentation, illustrated in Figure \ref{fig:architecture2}, is based on the cycle-consistent GAN (CycleGAN) model \cite{zhu2017unpaired} which has shown outstanding performance for unpaired image-to-image translation. This architecture is composed of four inter-connected networks, two conditional generators and two discriminators, which are trained simultaneously. In the original CycleGAN model, the generators are employed to learn a bidirectional mapping from an image domain to the other. On the other hand, discriminators try to determine whether an image from the corresponding domain is real or generated. By fooling the discriminators through adversarial learning, the model thus learns to generate images from the true distribution without requiring paired images. A cycle-consistency loss is also added to ensure that generators are consistent, i.e. that we recover the same image when going through both generators sequentially. 

\begin{figure*}[ht!]
\begin{center}
 \setlength{\tabcolsep}{1pt}
 \begin{tabular}{cccc}
 \includegraphics[width=0.218\linewidth,trim={0 12mm 0 12mm},clip]{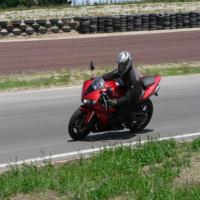} & 
 \includegraphics[width=0.218\linewidth,trim={0 12mm 0 12mm},clip]{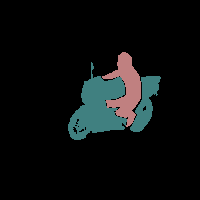} &
 \includegraphics[width=0.218\linewidth,trim={0 12mm 0 12mm},clip]{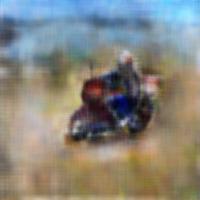} &
 \includegraphics[width=0.218\linewidth,trim={0 12mm 0 12mm},clip]{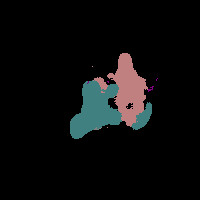} \\
 \includegraphics[width=0.218\linewidth]{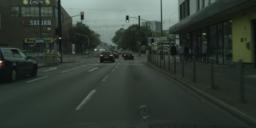} & 
 \includegraphics[width=0.218\linewidth]{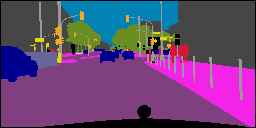} &
 \includegraphics[width=0.218\linewidth]{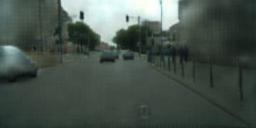} &
 \includegraphics[width=0.218\linewidth]{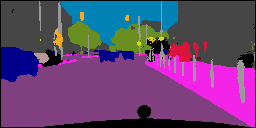} \\
 \includegraphics[width=0.218\linewidth,trim={0 14mm 0 10mm},clip]{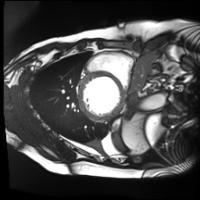} & 
 \includegraphics[width=0.218\linewidth,trim={0 14mm 0 10mm},clip]{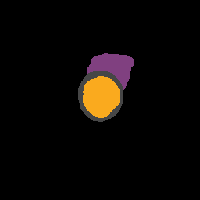} &
 \includegraphics[width=0.218\linewidth,trim={0 14mm 0 10mm},clip]{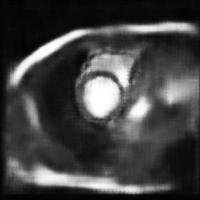} &
 \includegraphics[width=0.218\linewidth,trim={0 14mm 0 10mm},clip]{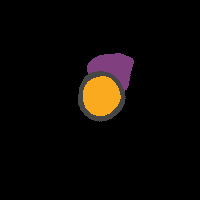} \\
 Image & Ground truth labels & Generated image & Generated labels
 \end{tabular}
 \end{center}
 \caption{Examples of images, ground truth labels, generated images and generated labels obtained for three benchmark datasets: PASCAL VOC 2012 (top row), Cityscapes (middle row), and ACDC (bottom row).}
 \label{fig:illustration}
\end{figure*}

In our semi-supervised segmentation model, the CycleGAN is instead used to map images to their corresponding segmentation mask and vice-versa. The first generator ($\GenItoS$), corresponding to the segmentation network that we want to obtain, learns a mapping from an image to its segmentation labels. The first discriminator ($D_S$) tries to differentiate these generated labels from real segmentation masks. Note that the combination of $\GenItoS$ of $D_S$ is similar to semi-supervised segmentation approach presented in \cite{zhang2017deep}. Conversely, the second generator ($\GenStoI$) learns to map a segmentation mask to its image. In our semi-supervised segmentation setting, this generator is only used to improve training. Likewise, the second discriminator ($D_I$) receives an image as input and predicts whether this image is real or generated. To enforce cycle consistency, generators are trained so that feeding the labels generated by $\GenItoS$ for an image into $\GenStoI$ gives that same image, and passing back to $\GenItoS$ the image generated by $\GenStoI$ for a segmentation mask gives that same mask. Figure \ref{fig:illustration} shows examples of images, ground truth labels, generated images and generated labels obtained for the three datasets used in our experiments.

\subsection{Loss functions}

In this section, we formally define the loss functions employed to train our model in a semi-supervised setting where the data comes from three distributions: labeled images ($\DistrImgLab$), ground truth masks of labeled images ($\DistrSegLab$), and unlabeled images ($\DistrImgUnlab$). The first loss function is a standard supervised segmentation loss that imposes the segmentation network ($\GenItoS$) to generate labels of ground truth masks:
\begin{equation}
\begin{split}
\LossCrossEntropy(\GenItoS) \ = \ \E_{\ImgLab, \SegLab \sim \DistrImgLab, \DistrSegLab} \big[ \Xentr\big(\SegLab, \GenItoS(\ImgLab)\big) \big]
\end{split}
\end{equation}
where $\Xentr$ is the pixelwise cross-entropy defined as
\begin{equation}
\Xentr(\SegLab, \widehat{\SegLab})  \ = \ - \sum_{j=1}^N \sum_{k=1}^K \SegLab_{j,k} \log \, \widehat{\SegLab}_{j,k}.
\end{equation}
In this expression, $\SegLab_{j,k}$ and $\widehat{\SegLab}_{j,k}$ are the ground truth and predicted probabilities that pixel $j \in \{1, \ldots, N\}$ has label $k \in \{1, \ldots, K\}$. Likewise, we employ a pixelwise L2 norm between a labeled image and the image generated from its corresponding ground truth as supervised loss to train the image generator $\GenStoI$:
\begin{equation}
\LossNorm(\GenStoI) \ = \ \E_{\ImgLab, \SegLab \sim \DistrImgLab, \DistrSegLab} \big[ \left \| \GenStoI(\SegLab) - \ImgLab \right \|^2_2 \big].
\end{equation}

To exploit unlabeled images, we incorporate two additional types of losses: adversarial losses and cycle-consistency losses. The adversarial losses are used to train the generators and discriminators in a competing fashion, and help the generators produce realistic images and segmentation masks. To have a better training of the discriminators, we follow the approach presented in \cite{mao2017least} and use a least square loss instead of the traditional cross-entropy. It was shown in this previous work that this loss function leads to minimizing the Pearson $\chi^2$ divergence. Suppose that $D_S(\SegLab)$ is the predicted probability that segmentation labels $\SegLab$ correspond to a ground truth mask. We define the adversarial loss for $D_S$ as
\begin{equation}
\begin{split}
    \LossUnlabAdv^S(\GenItoS, D_S)& \ = \ \E_{\SegLab \sim \DistrSegLab} \big[\big(D_S(\SegLab)-1\big)^2 \big] \, + \\
& \E_{\ImgUnlab \sim \DistrImgUnlab} \big[\big( D_S(\GenItoS(\ImgUnlab))\big)^2 \big].
\end{split}
\end{equation}
Similarly, let $D_I(x)$ be the predicted probability that an image $x$ is real, the adversarial loss for the image discriminator is defined as
\begin{equation}
\begin{split}
    \LossUnlabAdv^I(\GenStoI, D_I)& \ = \ \E_{\ImgUnlab \sim \DistrImgUnlab} \big[\big(D_I(\ImgUnlab) -1 \big)^2 \big] \, + \\
& \E_{\SegLab \sim \DistrSegLab} \big[\big(D_I(\GenStoI(\SegLab))\big)^2\big].
\end{split}
\end{equation}

The first cycle consistency loss measures the difference between an unlabeled image and the regenerated image after passing through generators $G_{IS}$ and $G_{SI}$ sequentially. Here, we use the L1 norm since it leads to sharper images than the L2 norm:
\begin{equation}
 \LossCycle^I(\GenItoS,\GenStoI) \ = \ \E_{\ImgUnlab \sim \DistrImgUnlab} \big[\left \|  \GenStoI(\GenItoS(\ImgUnlab)) - \ImgUnlab \right \|_1 \big].
\end{equation}
On the other hand, since the segmentation labels are categorical variables, we use cross-entropy to evaluate the difference between a ground-truth segmentation mask and the regenerated labels after passing through generators $G_{SI}$ and $G_{IS}$ in sequence:
\begin{equation}
 \LossCycle^S(\GenItoS,\GenStoI) \ = \ \E_{\SegLab \sim \DistrSegLab} \big[\Xentr\big(\SegLab, \GenItoS(\GenStoI(\SegLab))\big)\big].   
\end{equation}
Finally, the total loss is obtained by combining all six loss terms:
\begin{equation}
\begin{split}
& \LossTotal(\GenItoS, \GenStoI, D_S, D_I) \ = \ \LossCrossEntropy(\GenItoS) \, + \, \lambda_1 \LossNorm(\GenStoI) \\
    & \qquad + \, \lambda_2 \LossCycle^S(\GenItoS, \GenStoI) \, + \, \lambda_3 \LossCycle^I(\GenItoS, \GenStoI) \\
    & \qquad\qquad - \,\lambda_4\LossUnlabAdv^S(\GenItoS,D_S) \, - \, 
        \lambda_5\LossUnlabAdv^I(\GenStoI,D_I)
\end{split}
\label{eq:totalloss}
\end{equation}
\begin{equation}
    \argmin_{\GenItoS, \GenStoI} \ \argmax_{D_S,D_I} \ \LossTotal(\GenItoS, \GenStoI, D_S, D_I).
\end{equation}
In practice, learning is performed in an alternating fashion, where the parameters of the generators are optimized while considering those of the discriminators as fixed, and vice-versa.

\subsection{Implementation details} 

Following the original implementation of CycleGAN, we adopt the architecture proposed in \cite{johnson2016perceptual} for our generators, since it has shown impressive results for image-style transfer. This network is composed of two stride-2 convolutions, followed by 9 residual blocks and two fractionally-strided convolutions with stride $\nicefrac{1}{2}$. Similarly, instance normalization \cite{ulyanov2016instance} was employed and no drop-out was adopted. Furthermore, we used softmax as output function when generating segmentation labels from images, whereas \textit{tanh} was the selected function when translating from segmentation labels to images, in order to have continuous values. In pre-processing, each channel of an image is normalized to the $[-1,1]$ range by subtracting its mean value and dividing by the difference between the maximum and minimum value. 

Unlike the original CycleGAN model, we make use of pixel-wise discriminators \cite{Hung_semiseg_2018} where the size of the output is the same as the input and the adversarial label (i.e., real\,/\,generated) is recopied at each output pixel. We found this model to perform better than having a single discriminator output. Each discriminator contains three convolutional blocks, followed by Leaky ReLU activations with negative slope of $\alpha\!=\!0.2$. In addition, batch normalization is used in the discriminators after the second convolutional block.

Both generators and discriminators were trained using Adam optimizer \cite{kingma2014adam} with $\beta_1$ and $\beta_2$ parameters equal to 0.5 and 0.999. Learning rate was initially set to 2$\times$10$^{-4}$ with a linear decay after every 100 epochs, during 400 epochs. Furthermore, batch size was set to 5 in all experiments. The values of the weighting terms in Eq. (\ref{eq:totalloss}) were set to $\lambda_1\!=\!1$, $\lambda_2\!=\!0.05$, $\lambda_3\!=\!1$, $\lambda_4\!=\!1$ and $\lambda_5\!=\!1$. The code was implemented in Pytorch 3.3 \cite{paszke2017automatic} and experiments were run on a server equipped with a NVIDIA Titan V GPU (12 GBs). The code is made publicly available at \url{https://github.com/arnab39/Semi-supervised-segmentation-cycleGAN}. 

\section{Experiments}\label{sec:experiments}

\subsection{Datasets}
\label{ssec:dataset}

We conduct experiments on three different public semantic segmentation benchmarks: PASCAL VOC 2012 \cite{everingham2010pascal}, Cityscapes \cite{cordts2016cityscapes} and the Automated Cardiac Diagnosis Challenge (ACDC) MICCAI 2017 Challenge \cite{bernard2018deep}.\\[2mm] 
\noindent\textbf{PASCAL VOC 2012:} This dataset contains 21 common object classes, including one background class. In our experiments, we employed the augmented set composed of 10,582 images, which we split into training (8994 images) and validation (1588 images) subsets. In addition, due to memory limitations, we resized all images to 200$\times$200 pixels before being fed into the network. \\[2mm]
\noindent\textbf{Cityscapes:} This second dataset contains 50 videos from driving scenes where a total of 20 classes (including background) are manually annotated. In our experiments, we split the 3475 provided images into training (2953 images) and validation (522 images) subsets. As in the previous case, all images were resized to a 128$\times$256 pixel resolution.\\[2mm]
\noindent\textbf{ACDC:} This medical image set focuses on the segmentation of cardiac structures (the left ventricular endocardium and epicardium and the right ventricular endocardium) and consists of 100 cine magnetic resonance (MR) exams covering normal cases and subjects with well-defined defined pathologies: dilated cardiomyopathy, hypertrophic cardiomyopathy, myocardial infarction with altered left ventricular ejection fraction and abnormal right ventricle. Each exam contains acquisitions at the diastolic and systolic phases. For our experiments, we employed 75 exams for training and the remaining 25 for validation. 

It is important to note that, since we aim at isolating the performance of each method and not achieving state-of-the-art results, no data augmentation was performed in any of the datasets for training.


\subsection{Evaluation protocol} 


We use the mean intersection over union (mIOU) metric to evaluate the segmentation results of all the models. This metric can be defined as $\frac{\mr{TP}}{\mr{TP}+\mr{FP}+\mr{FN}}$, where $\mr{TP}$, $\mr{FP}$, and $\mr{FN}$ are the true positive, false positive, and false negative pixels, respectively, determined over the whole validation set. 



To have an upper-bound on performance, we train a network in a fully-supervised manner, employing all available training images. We also trained the same model from scratch using only 10\%, 20\%, 30\% or 50\% of labeled images, and refer to this baseline as Partial. Our semi-supervised method is trained with the same subsets as the Partial baseline, however it also makes use of unlabeled training images. Last, we compare our method to the approach presented in \cite{Hung_semiseg_2018}, which has shown state-of-art performance for semi-supervised segmentation.

\subsection{Results}

In the following section, we report the experimental results of the proposed approach on the three datasets described in Section \ref{ssec:dataset}.

\subsubsection{Comparison on benchmarks}

\begin{table}[ht!]
\centering
\begin{small}\setlength{\tabcolsep}{4pt}
\begin{tabular}{lcccc}
\toprule
\textbf{Method} & \textbf{Labeled\,\%} & \textbf{VOC} & \textbf{Cityscapes} & \textbf{ACDC} \\
\midrule\midrule
Full & 100 & 0.5543 & 0.5551 & 0.8982 \\
\midrule
Hung et al. \cite{Hung_semiseg_2018} & 20 & 0.2032 & 0.3490 & 0.8063 \\
\midrule
\multirow{4}{*}{Partial} & 50 &  0.4108 & 0.4856 & 0.8863 \\
& 30 &  0.3237 & 0.4502 & 0.8785 \\
& 20 &  0.2688 & 0.4112 & 0.8642 \\
& 10 &  0.2158 & 0.3636 & 0.8418 \\
\midrule
\multirow{4}{*}{Ours} & 50 &  0.4197 & 0.4997 & 0.8890 \\
& 30 &  0.3514 & 0.4654 & 0.8804 \\
& 20 &  0.2981 & 0.4321 & 0.8688 \\
& 10 &  0.2543 & 0.3923 & 0.8463 \\
\bottomrule
\end{tabular}
\end{small}
\caption{Semantic segmentation performance of tested methods on the three benchmark datasets, for different levels of supervision. \emph{Full} corresponds to the segmentation network trained with all training samples, and \emph{Partial} to the same network trained with a subset of labeled images (ranging from 10\% to 50\%) without considering unlabeled images.}
\label{tab:quantitative_All}
\end{table}

Table \ref{tab:quantitative_All} reports the results obtained by the tested approaches on the three benchmark datasets. We first observe that, in all cases, the proposed model outperforms the partial supervision baseline when training with a reduced set of labeled images. This difference is particularly significant when pixel-level annotations are scarce (i.e., 10\% and 20\% of the whole training set), where the proposed model achieves 2-4\% of improvement. As the number of labeled images increases, the gap between the baseline and the proposed models decreases, with a gain close to 1\% when training with half of the whole training set. Furthermore, we found that the semi-supervised segmentation approach of Hung et al. \cite{Hung_semiseg_2018} obtained poor results for all three datasets, with lower accuracy than the partial supervision baseline (Partial). In the original work \cite{Hung_semiseg_2018}, authors used a generator pre-trained using ImageNet. In our experiments, to have an unbiased comparison, we tested methods without such pre-training (i.e., all generators and discriminators were trained from scratch). This could potentially explain our lower results obtained for Hung et al.'s method.   
\begin{figure*}[ht!]
\begin{center}
 \setlength{\tabcolsep}{1pt}
 \begin{tabular}{cccccc}
 \visVOC{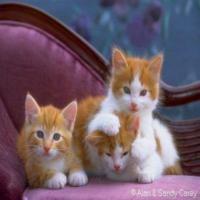} & 
 \visVOC{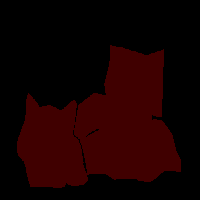} &
 \visVOC{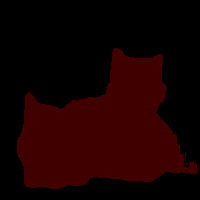} &
 \visVOC{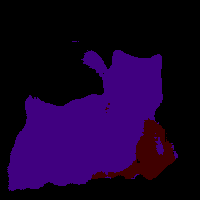} &
 \visVOC{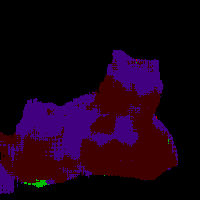} &
 \visVOC{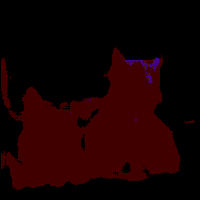} \\
 \visVOC{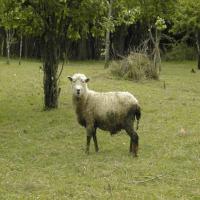} & 
 \visVOC{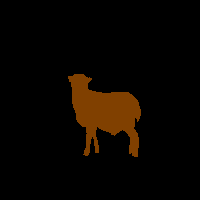} &
 \visVOC{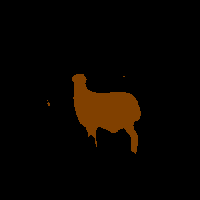} &
 \visVOC{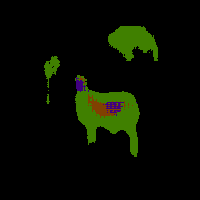} &
 \visVOC{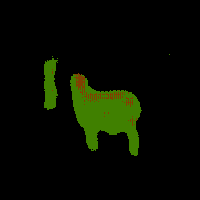} &
 \visVOC{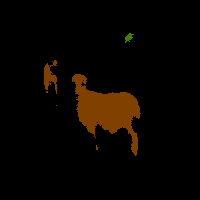} \\
 \visVOC{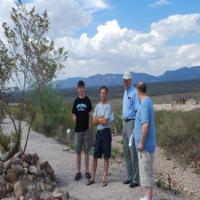} & 
 \visVOC{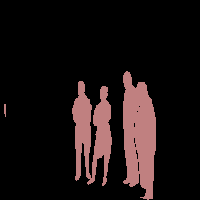} &
 \visVOC{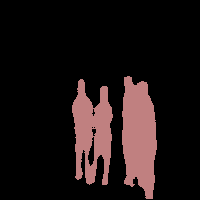} &
 \visVOC{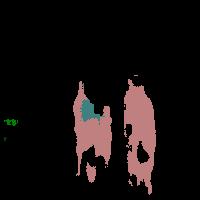} &
 \visVOC{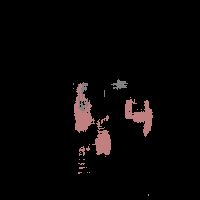} &
 \visVOC{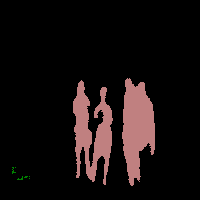} \\
 Image & Ground truth & Full & Partial & Hung et al. \cite{Hung_semiseg_2018} & Ours
 \end{tabular}
 \end{center}
 \vspace*{-4mm}
 \caption{Visual comparisons on the PASCAL VOC 2012 dataset employing 20\% of labeled images for training.}
 \label{fig:illustrationVOC}
\end{figure*}
 
 \begin{figure*}[ht!]
\begin{center}
 \setlength{\tabcolsep}{1pt}
 \begin{tabular}{ccc}
 \visCity{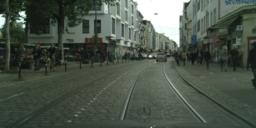} & 
 \visCity{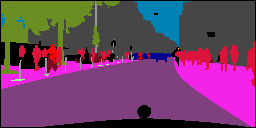} &
 \visCity{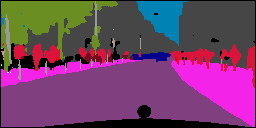} \\[-.5mm]
 Image & Ground truth & Full \\
 \visCity{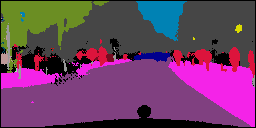} &
 \visCity{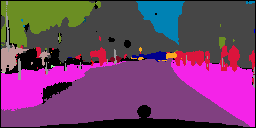} &
 \visCity{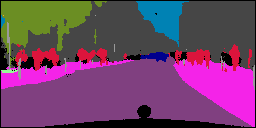}  \\[-.5mm]
 Partial & Hung et al. \cite{Hung_semiseg_2018} & Ours \\[3mm] 
 \visCity{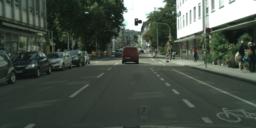} & 
 \visCity{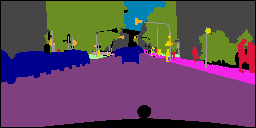} &
 \visCity{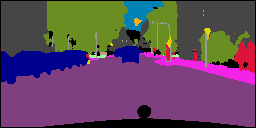} \\[-.5mm]
  Image & Ground truth & Full \\
 \visCity{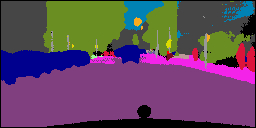} &
 \visCity{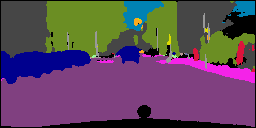} &
 \visCity{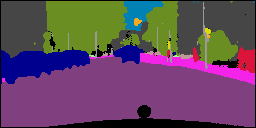}  \\[-.5mm]
 Partial & Hung et al. \cite{Hung_semiseg_2018} & Ours 
 \end{tabular}
 \end{center}
 \vspace*{-4mm}
 \caption{Visual comparisons on the Cityscape dataset employing 20\% of labeled images for training.}
 \label{fig:illustrationCityscapes}
\end{figure*}

 \begin{figure*}[ht!]
\begin{center}
 \setlength{\tabcolsep}{1pt}
 \begin{tabular}{cccccc}
 \visACDC{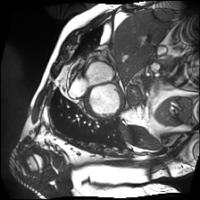} & 
 \visACDC{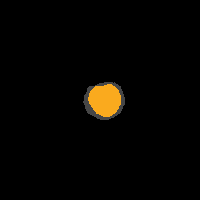} &
 \visACDC{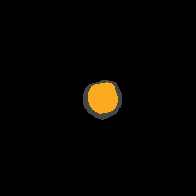} &
 \visACDC{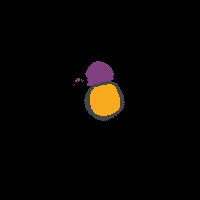} &
 \visACDC{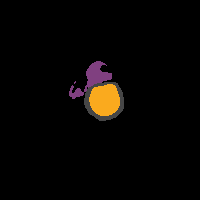} &
 \visACDC{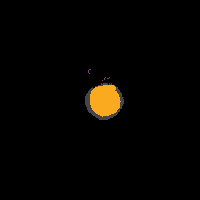} \\
 \visACDC{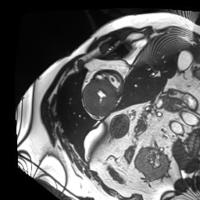} & 
 \visACDC{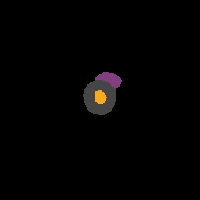} &
 \visACDC{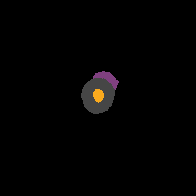} &
 \visACDC{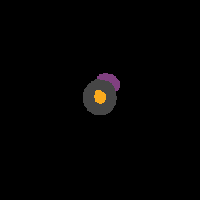} &
 \visACDC{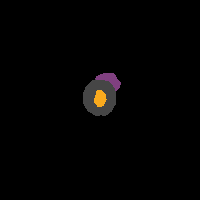} &
 \visACDC{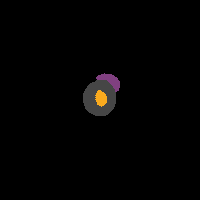} \\
 Image & Ground truth & Full & Partial & Hung et al. \cite{Hung_semiseg_2018} & Ours
 \end{tabular}
 \end{center}
 \vspace*{-4mm}
 \caption{Visual comparisons on the ACDC dataset employing 20\% of labeled images for training.}
 \label{fig:illustrationACDC}
\end{figure*}

A visual comparison of results is given in Figures \ref{fig:illustrationVOC}, \ref{fig:illustrationCityscapes} and \ref{fig:illustrationACDC}. It can be seen that the proposed method predicts a segmentation closer to the network trained with all images (Full) than the partial supervision baseline (Partial) and the Hung et al.'s model. While predicted region boundaries are sometimes inaccurate, the global semantic information of the image (i.e., actual class labels) appears to be better learned by our model compared to the partial supervision baseline. In addition, our model seems to better capture details of thin objects --e.g., legs or persons-- compared to both the baseline and the method in \cite{Hung_semiseg_2018}.    

\begin{table}[ht!]
\centering
\begin{tabular}{lcccc}
\toprule
\textbf{Method} & \textbf{VOC}  \\
\midrule\midrule
Proposed & 0.2981 \\
\midrule
w/o labels cycle loss $\big(\LossCycle^S\big)$ & 0.2627 \\
w/o image cycle loss $\big(\LossCycle^I\big)$ & 0.2733 \\
w/o labels discr. loss $\big(\LossUnlabAdv^S\big)$ & 0.2614 \\
w/o image discr. loss $\big(\LossUnlabAdv^I\big)$ & 0.2543 \\
\bottomrule
\end{tabular}
\caption{Ablation study on the PASCAL VOC 2012 dataset with 20\% labeled data.}
\label{table:ablation}
\end{table}

\subsubsection{Ablation study}

To further analyze the effect of the different components of the proposed model, we conduct an ablation study where the model is trained while removing a single loss term of Eq. (\ref{eq:totalloss}). Specifically, we train the model without the labels cycle-consistency loss $\big(\LossCycle^S\big)$, image cycle-consistency loss $\big(\LossCycle^S\big)$, labels discriminator loss $\big(\LossUnlabAdv^S\big)$, or image discriminator loss $\big(\LossUnlabAdv^I\big)$. Note that these modifications correspond to setting $\lambda_3$, $\lambda_4$, $\lambda_5$ or $\lambda_6$ to 0, respectively. For this experiment, we investigate the performance of the model trained with 20\% of labeled data on PASCAL VOC 2012.

The results of our ablation study are summarized in Table \ref{table:ablation}. The proposed model containing all loss terms reaches a mIOU value of 0.2981. If we remove the cycle consistency loss on the generation of segmentation labels, this value is reduced to 0.2627. However, removing the cycle consistency loss on the image generation leads to an even lower accuracy of 0.2733, suggesting that the cycle consistency loss on segmentation masks has a stronger impact in the model. Regarding the significance of the losses in the discriminators, we observe a reverse effect. A lower performance is observed if the loss on the discriminator $D_I$ is ignored, which is responsible of differentiating between unlabeled and generated images.

\section{Discussion and conclusion}



We presented a semi-supervised method for image semantic segmentation, where the key idea is to leverage CycleGAN to learn a cycle-consistent mapping between unlabeled real images and available ground truth masks. Unlike recent work using adversarial learning for semi-supervised segmentation \cite{souly2017semi,Hung_semiseg_2018,zhang2017deep}, the proposed strategy enforces consistency between unpaired images and segmentation masks, which acts as an unsupervised regularizer. From the reported results, we have shown that this strategy improves segmentation performance, particularly when annotated data is scarce.

Due to the high computational and memory requirements of generating large images, our experiments have employed images with reduced size, in particular for the Cityscape dataset where the resolution was reduced from $1024\!\times\!2048$ pixels to $128\!\times\!256$. This is in large part responsible for the lower accuracy values obtained in our experiments, compared to those reported in the literature. In a future investigation, we will evaluate the performance of our model on full-sized images. Moreover, in this work, we used the same network for both generators ($\GenItoS$ and $\GenStoI$). This architectural choice was made to achieve a better learning equilibrium during training (i.e., avoid a generator learning much faster than the other). Employing different networks in future experiments could however improve performance. 


{\small
\bibliographystyle{ieee}
\bibliography{egbib}

\begin{thebibliography}{1}\itemsep=-1pt

\bibitem{Alpher02}
A.~Alpher.
\newblock Frobnication.
\newblock {\em Journal of Foo}, 12(1):234--778, 2002.

\bibitem{Alpher03}
A.~Alpher and J.~P.~N. Fotheringham-Smythe.
\newblock Frobnication revisited.
\newblock {\em Journal of Foo}, 13(1):234--778, 2003.

\bibitem{Alpher04}
A.~Alpher, J.~P.~N. Fotheringham-Smythe, and G.~Gamow.
\newblock Can a machine frobnicate?
\newblock {\em Journal of Foo}, 14(1):234--778, 2004.

\bibitem{Authors06b}
Authors.
\newblock Frobnication tutorial, 2006.
\newblock Supplied as additional material {\tt tr.pdf}.

\bibitem{Authors06}
Authors.
\newblock The frobnicatable foo filter, 2011.
\newblock Face and Gesture submission ID 324. Supplied as additional material
  {\tt fg324.pdf}.

\end{thebibliography}


\begin{thebibliography}{10}\itemsep=-1pt

\bibitem{bai2017semi}
W.~Bai, O.~Oktay, M.~Sinclair, H.~Suzuki, M.~Rajchl, G.~Tarroni, B.~Glocker,
  A.~King, P.~M. Matthews, and D.~Rueckert.
\newblock Semi-supervised learning for network-based cardiac mr image
  segmentation.
\newblock In {\em International Conference on Medical Image Computing and
  Computer-Assisted Intervention}, pages 253--260. Springer, 2017.

\bibitem{baur2017semi}
C.~Baur, S.~Albarqouni, and N.~Navab.
\newblock Semi-supervised deep learning for fully convolutional networks.
\newblock In {\em International Conference on Medical Image Computing and
  Computer-Assisted Intervention}, pages 311--319. Springer, 2017.

\bibitem{bernard2018deep}
O.~Bernard, A.~Lalande, C.~Zotti, F.~Cervenansky, X.~Yang, P.-A. Heng,
  I.~Cetin, K.~Lekadir, O.~Camara, M.~A.~G. Ballester, et~al.
\newblock Deep learning techniques for automatic {MRI} cardiac multi-structures
  segmentation and diagnosis: Is the problem solved?
\newblock {\em IEEE Transactions on Medical Imaging}, 2018.

\bibitem{Chen2018DeepLab}
L.~C. Chen, G.~Papandreou, I.~Kokkinos, K.~Murphy, and A.~L. Yuille.
\newblock Deeplab: Semantic image segmentation with deep convolutional nets,
  atrous convolution, and fully connected crfs.
\newblock {\em IEEE Transactions on Pattern Analysis and Machine Intelligence},
  40(4):834--848, 2018.

\bibitem{cordts2016cityscapes}
M.~Cordts, M.~Omran, S.~Ramos, T.~Rehfeld, M.~Enzweiler, R.~Benenson,
  U.~Franke, S.~Roth, and B.~Schiele.
\newblock The cityscapes dataset for semantic urban scene understanding.
\newblock In {\em Proceedings of the IEEE conference on computer vision and
  pattern recognition}, pages 3213--3223, 2016.

\bibitem{dai2015boxsup}
J.~Dai, K.~He, and J.~Sun.
\newblock Boxsup: Exploiting bounding boxes to supervise convolutional networks
  for semantic segmentation.
\newblock In {\em Proceedings of the IEEE International Conference on Computer
  Vision}, pages 1635--1643, 2015.

\bibitem{everingham2010pascal}
M.~Everingham, L.~Van~Gool, C.~K. Williams, J.~Winn, and A.~Zisserman.
\newblock The pascal visual object classes (voc) challenge.
\newblock {\em International journal of computer vision}, 88(2):303--338, 2010.

\bibitem{NIPS2014_5423}
I.~Goodfellow, J.~Pouget-Abadie, M.~Mirza, B.~Xu, D.~Warde-Farley, S.~Ozair,
  A.~Courville, and Y.~Bengio.
\newblock Generative adversarial nets.
\newblock In Z.~Ghahramani, M.~Welling, C.~Cortes, N.~D. Lawrence, and K.~Q.
  Weinberger, editors, {\em Advances in Neural Information Processing Systems
  27}, pages 2672--2680. Curran Associates, Inc., 2014.

\bibitem{hoffman2017cycada}
J.~Hoffman, E.~Tzeng, T.~Park, J.-Y. Zhu, P.~Isola, K.~Saenko, A.~A. Efros, and
  T.~Darrell.
\newblock Cycada: Cycle-consistent adversarial domain adaptation.
\newblock {\em arXiv preprint arXiv:1711.03213}, 2017.

\bibitem{hoffman2016fcns}
J.~Hoffman, D.~Wang, F.~Yu, and T.~Darrell.
\newblock {FCN}s in the wild: Pixel-level adversarial and constraint-based
  adaptation.
\newblock {\em arXiv preprint arXiv:1612.02649}, 2016.

\bibitem{Hung_semiseg_2018}
W.-C. Hung, Y.-H. Tsai, Y.-T. Liou, Y.-Y. Lin, and M.-H. Yang.
\newblock Adversarial learning for semi-supervised semantic segmentation.
\newblock In {\em Proceedings of the British Machine Vision Conference (BMVC)},
  page~1, 2018.

\bibitem{isola2017image}
P.~Isola, J.-Y. Zhu, T.~Zhou, and A.~A. Efros.
\newblock Image-to-image translation with conditional adversarial networks.
\newblock In {\em Proceedings of the IEEE conference on computer vision and
  pattern recognition}, pages 1125--1134, 2017.

\bibitem{jiang2018tumor}
J.~Jiang, Y.-C. Hu, N.~Tyagi, P.~Zhang, A.~Rimner, G.~S. Mageras, J.~O. Deasy,
  and H.~Veeraraghavan.
\newblock Tumor-aware, adversarial domain adaptation from ct to mri for lung
  cancer segmentation.
\newblock In {\em International Conference on Medical Image Computing and
  Computer-Assisted Intervention}, pages 777--785. Springer, 2018.

\bibitem{johnson2016perceptual}
J.~Johnson, A.~Alahi, and L.~Fei-Fei.
\newblock Perceptual losses for real-time style transfer and super-resolution.
\newblock In {\em European conference on computer vision}, pages 694--711.
  Springer, 2016.

\bibitem{kervadec2019constrained}
H.~Kervadec, J.~Dolz, M.~Tang, E.~Granger, Y.~Boykov, and I.~B. Ayed.
\newblock Constrained-{CNN} losses for weakly supervised segmentation.
\newblock {\em Medical image analysis}, 2019.

\bibitem{kingma2014adam}
D.~P. Kingma and J.~Ba.
\newblock Adam: A method for stochastic optimization.
\newblock {\em arXiv preprint arXiv:1412.6980}, 2014.

\bibitem{lin2016scribblesup}
D.~Lin, J.~Dai, J.~Jia, K.~He, and J.~Sun.
\newblock Scribblesup: Scribble-supervised convolutional networks for semantic
  segmentation.
\newblock In {\em Proceedings of the IEEE Conference on Computer Vision and
  Pattern Recognition}, pages 3159--3167, 2016.

\bibitem{long2015fully}
J.~Long, E.~Shelhamer, and T.~Darrell.
\newblock Fully convolutional networks for semantic segmentation.
\newblock In {\em Proceedings of the IEEE conference on computer vision and
  pattern recognition}, pages 3431--3440, 2015.

\bibitem{mao2017least}
X.~Mao, Q.~Li, H.~Xie, R.~Y. Lau, Z.~Wang, and S.~Paul~Smolley.
\newblock Least squares generative adversarial networks.
\newblock In {\em Proceedings of the IEEE International Conference on Computer
  Vision}, pages 2794--2802, 2017.

\bibitem{min2018robust}
S.~Min and X.~Chen.
\newblock A robust deep attention network to noisy labels in semi-supervised
  biomedical segmentation.
\newblock {\em arXiv preprint arXiv:1807.11719}, 2018.

\bibitem{papandreou2015weakly}
G.~Papandreou, L.-C. Chen, K.~Murphy, and A.~L. Yuille.
\newblock Weakly-and semi-supervised learning of a {DCNN} for semantic image
  segmentation.
\newblock In {\em ICCV}, 2015.

\bibitem{paszke2017automatic}
A.~Paszke, S.~Gross, S.~Chintala, G.~Chanan, E.~Yang, Z.~DeVito, Z.~Lin,
  A.~Desmaison, L.~Antiga, and A.~Lerer.
\newblock Automatic differentiation in {PyTorch}.
\newblock In {\em NIPS Autodiff Workshop}, 2017.

\bibitem{peng2019deep}
J.~Peng, G.~Estradab, M.~Pedersoli, and C.~Desrosiers.
\newblock Deep co-training for semi-supervised image segmentation.
\newblock {\em arXiv preprint arXiv:1903.11233}, 2019.

\bibitem{perone2018unsupervised}
C.~S. Perone, P.~Ballester, R.~C. Barros, and J.~Cohen-Adad.
\newblock Unsupervised domain adaptation for medical imaging segmentation with
  self-ensembling.
\newblock {\em arXiv preprint arXiv:1811.06042}, 2018.

\bibitem{rajchl2017deepcut}
M.~Rajchl, M.~C. Lee, O.~Oktay, K.~Kamnitsas, J.~Passerat-Palmbach, W.~Bai,
  M.~Damodaram, M.~A. Rutherford, J.~V. Hajnal, B.~Kainz, et~al.
\newblock Deepcut: Object segmentation from bounding box annotations using
  convolutional neural networks.
\newblock {\em IEEE transactions on medical imaging}, 36(2):674--683, 2017.

\bibitem{ronneberger2015u}
O.~Ronneberger, P.~Fischer, and T.~Brox.
\newblock U-net: Convolutional networks for biomedical image segmentation.
\newblock In {\em International Conference on Medical image computing and
  computer-assisted intervention}, pages 234--241. Springer, 2015.

\bibitem{souly2017semi}
N.~Souly, C.~Spampinato, and M.~Shah.
\newblock Semi supervised semantic segmentation using generative adversarial
  network.
\newblock In {\em Computer Vision (ICCV), 2017 IEEE International Conference
  on}, pages 5689--5697. IEEE, 2017.

\bibitem{tsai2018learning}
Y.-H. Tsai, W.-C. Hung, S.~Schulter, K.~Sohn, M.-H. Yang, and M.~Chandraker.
\newblock Learning to adapt structured output space for semantic segmentation.
\newblock In {\em Computer Vision and Pattern Recognition (CVPR)}, 2018.

\bibitem{tzeng2017adversarial}
E.~Tzeng, J.~Hoffman, K.~Saenko, and T.~Darrell.
\newblock Adversarial discriminative domain adaptation.
\newblock In {\em Computer Vision and Pattern Recognition (CVPR)}, volume~1,
  page~4, 2017.

\bibitem{ulyanov2016instance}
D.~Ulyanov, A.~Vedaldi, and V.~Lempitsky.
\newblock Instance normalization: The missing ingredient for fast stylization.
\newblock {\em arXiv preprint arXiv:1607.08022}, 2016.

\bibitem{zhang2017deep}
Y.~Zhang, L.~Yang, J.~Chen, M.~Fredericksen, D.~P. Hughes, and D.~Z. Chen.
\newblock Deep adversarial networks for biomedical image segmentation utilizing
  unannotated images.
\newblock In {\em International Conference on Medical Image Computing and
  Computer-Assisted Intervention}, pages 408--416, 2017.

\bibitem{zhou2019collaborative}
Y.~Zhou, X.~He, L.~Huang, L.~Liu, F.~Zhu, S.~Cui, and L.~Shao.
\newblock Collaborative learning of semi-supervised segmentation and
  classification for medical images.
\newblock In {\em Proceedings of the IEEE Conference on Computer Vision and
  Pattern Recognition}, pages 2079--2088, 2019.

\bibitem{zhou2018semi}
Y.~Zhou, Y.~Wang, P.~Tang, W.~Shen, E.~K. Fishman, and A.~L. Yuille.
\newblock Semi-supervised multi-organ segmentation via multi-planar
  co-training.
\newblock {\em arXiv preprint arXiv:1804.02586}, 2018.

\bibitem{zhu2017unpaired}
J.-Y. Zhu, T.~Park, P.~Isola, and A.~A. Efros.
\newblock Unpaired image-to-image translation using cycle-consistent
  adversarial networks.
\newblock In {\em Proceedings of the IEEE International Conference on Computer
  Vision}, pages 2223--2232, 2017.

\end{thebibliography}
}

\begin{table*}
\begin{center}
\begin{tabular}{cccccccc}
\toprule
 & \multicolumn{3}{c}{\textbf{Partial}} & & \multicolumn{3}{c}{\textbf{Ours}} \\
\cmidrule(l{4pt}r{4pt}){2-4}\cmidrule(l{4pt}r{4pt}){6-8}
\textbf{Class~~} & 10\% & 20\% & 30\% & & 10\% & 20\% & 30\% \\
\midrule\midrule
1 & 0.4236 & 0.5302 & 0.5798 & & 0.4369 & 0.5808 & 0.5819 \\
2 & 0.2046 & 0.2624 & 0.3328 & & 0.1423 & 0.3574 & 0.4197 \\
3 & 0.1060 & 0.1542 & 0.1544 & & 0.0238 & 0.1511 & 0.2575 \\
4 & 0.1633 & 0.2479 & 0.2349 & & 0.1504 & 0.2048 & 0.3146 \\
5 & 0.0119 & 0.0535 & 0.0744 & & 0.0227 & 0.0188 & 0.0623 \\
6 & 0.5192 & 0.5310 & 0.6315 & & 0.6962 & 0.6867 & 0.6426 \\
7 & 0.3263 & 0.3368 & 0.4746 & & 0.4451 & 0.4877 & 0.5042 \\
8 & 0.2523 & 0.4050 & 0.3841 & & 0.3455 & 0.3123 & 0.4075 \\
9 & 0.1029 & 0.1226 & 0.1379 & & 0.2158 & 0.1429 & 0.1680 \\
10 & 0.0123 & 0.1888 & 0.1211 & & 0.1412 & 0.2021 & 0.1579 \\
11 & 0.0725 & 0.1868 & 0.3487 & & 0.0643 & 0.2097 & 0.2359 \\
12 & 0.2048 & 0.2496 & 0.2509 & & 0.2331 & 0.2302 & 0.2989 \\
13 & 0.0781 & 0.1033 & 0.1769 & & 0.2474 & 0.1056 & 0.2205 \\
14 & 0.3811 & 0.4126 & 0.5343 & & 0.1496 & 0.4601 & 0.5107 \\
15 & 0.5886 & 0.6528 & 0.6721 & & 0.5690 & 0.6992 & 0.6645 \\
16 & 0.0889 & 0.1230 & 0.0775 & & 0.1996 & 0.2179 & 0.1707 \\
17 & 0.1294 & 0.1061 & 0.2078 & & 0.1851 & 0.0488 & 0.2258 \\
18 & 0.1089 & 0.0654 & 0.1820 & & 0.2091 & 0.0350 & 0.2232 \\
19 & 0.2990 & 0.3760 & 0.4768 & & 0.1879 & 0.5115 & 0.4681 \\
20 & 0.2407 & 0.2668 & 0.4210 & & 0.4280 & 0.2988 & 0.5331 \\
\bottomrule
\end{tabular}
\caption{Classwise mean IoU for the 20 valid classes of the PASCAL VOC 2012 dataset, obtained with 10\%, 20\% or 30\% of labeled examples. \emph{Partial} corresponds to training only the segmentation network of our semi-supervised CycleGAN method with the subset of labeled examples.}
\label{tab:classwise_pascal}
\end{center}
\end{table*}

\begin{table*}
\centering
\begin{tabular}{cccccccc}
\toprule
 & \multicolumn{3}{c}{\textbf{Partial}} & & \multicolumn{3}{c}{\textbf{Ours}} \\
\cmidrule(l{4pt}r{4pt}){2-4}\cmidrule(l{4pt}r{4pt}){6-8}
\textbf{Class~~} & 10\% & 20\% & 30\% & & 10\% & 20\% & 30\% \\
\midrule\midrule
1 & 0.9346 & 0.9453 & 0.9523 & & 0.9369 & 0.9457 & 0.9534 \\ 
2 & 0.5932 & 0.6518 & 0.6850 & & 0.6323 & 0.6775 & 0.6998 \\ 
3 & 0.7807 & 0.8149 & 0.8333 & & 0.7977 & 0.8199 & 0.8489 \\ 
4 & 0.1419 & 0.1463 & 0.1820 & & 0.1401 & 0.1657 & 0.2008 \\ 
5 & 0.0715 & 0.1201 & 0.1684 & & 0.1128 & 0.1469 & 0.1726 \\ 
6 & 0.2441 & 0.2975 & 0.3288 & & 0.2732 & 0.3074 & 0.3426 \\ 
7 & 0.1255 & 0.2067 & 0.2481 & & 0.1869 & 0.2275 & 0.2644 \\ 
8 & 0.1993 & 0.2799 & 0.3315 & & 0.2528 & 0.3091 & 0.3512 \\ 
9 & 0.7824 & 0.8177 & 0.8355 & & 0.8012 & 0.8278 & 0.8565 \\ 
10 & 0.4458 & 0.4661 & 0.4923 & & 0.4433 & 0.4718 & 0.5137 \\ 
11 & 0.8777 & 0.8953 & 0.9019 & & 0.8872 & 0.8953 & 0.9072 \\ 
12 & 0.3766 & 0.4534 & 0.5021 & & 0.4387 & 0.4822 & 0.5229 \\ 
13 & 0.0687 & 0.1173 & 0.1786 & & 0.0867 & 0.1625 & 0.1961 \\ 
14 & 0.7626 & 0.8094 & 0.8367 & & 0.7822 & 0.8259 & 0.8528 \\ 
15 & 0.0877 & 0.0857 & 0.1258 & & 0.0825 & 0.1046 & 0.1473 \\ 
16 & 0.0448 & 0.0715 & 0.1735 & & 0.0576 & 0.1425 & 0.1965 \\ 
17 & 0.0814 & 0.2454 & 0.2892 & & 0.2067 & 0.2781 & 0.2985 \\ 
18 & 0.0739 & 0.0935 & 0.1471 & & 0.0701 & 0.1525 & 0.1529 \\ 
19 & 0.216 & 0.2948 & 0.3420 & & 0.2611 & 0.2601 & 0.3618 \\
\bottomrule
\end{tabular}
\caption{Classwise mean IoU for the 19 valid classes of the Cityscapes dataset, obtained with 10\%, 20\% or 30\% of labeled examples. \emph{Partial} corresponds to training only the segmentation network of our semi-supervised CycleGAN method with the subset of labeled examples.}
\label{tab:classwise_cityscapes}
\end{table*}

\newpage
\begin{figure*}
\begin{center}
 \setlength{\tabcolsep}{1pt}
 \begin{tabular}{cccc}
 \visVOCA{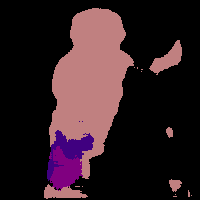}
 \visVOCA{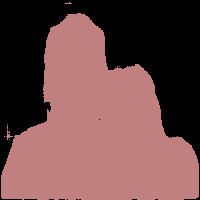}
 \visVOCA{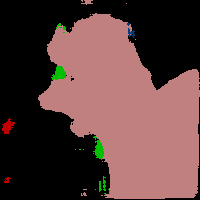}
 \visVOCA{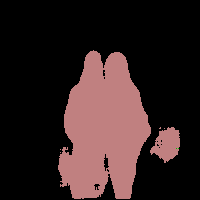}
 \visVOCA{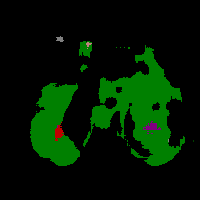}
 \visVOCA{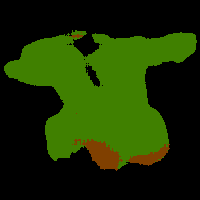}
 \visVOCA{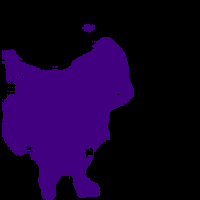}
 \visVOCA{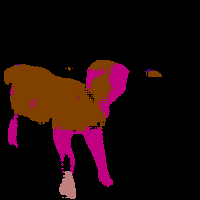}
 Image & Ground truth & Partial & Ours
 \end{tabular}
 \end{center}
 \vspace*{-4mm}
 \caption{Visual comparisons on the PASCAL VOC 2012 dataset employing 30\% of labeled images for training.}
 \label{fig:illustrationVOC}
\end{figure*}


\begin{figure*}
\begin{center}
 \setlength{\tabcolsep}{1pt}
 \begin{tabular}{cccc}
 \visCityA{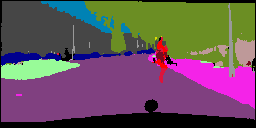}
 \visCityA{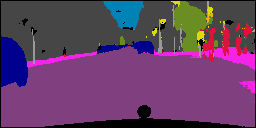}
 \visCityA{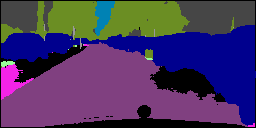}
 \visCityA{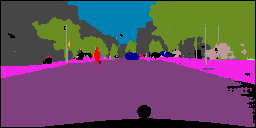}
 \visCityA{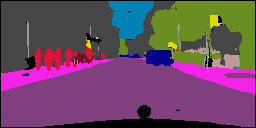}
 \visCityA{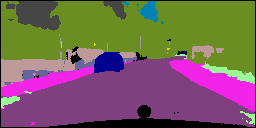}
 \visCityA{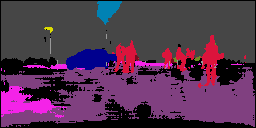}
 \visCityA{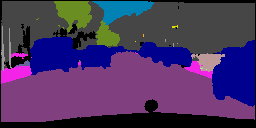}
 \visCityA{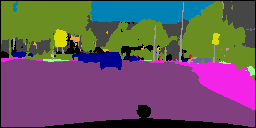}
 Image & Ground truth & Partial & Ours
 \end{tabular}
 \end{center}
 \vspace*{-4mm}
 \caption{Visual comparisons on the Cityscapes dataset employing 30\% of labeled images for training.}
 \label{fig:illustrationVOC}
\end{figure*}

\end{document}